\documentclass[fleqn,10pt,twocolumn]{SWARM25}

\title{Flocking Behavior: An Innovative Inspiration for the Optimization of Production Plants}

\author{M. Umlauft and M. Schranz}

\affils{Lakeside Labs GmbH, Lakeside B04b, 9020 Klagenfurt, Austria\\
(E-mail: \{umlauft, schranz\}@lakeside-labs.com)\\
}
\abstract{%
Optimizing modern production plants using the job-shop principle is a known hard problem. For very large plants, like semiconductor fabs, the problem becomes unsolvable on a plant-wide scale in a reasonable amount of time using classical linear optimization. An alternative approach is the use of swarm intelligence algorithms. These have been applied to the job-shop problem before, but often in a centrally calculated way where they are applied to the solution space, but they can be implemented in a bottom-up fashion to avoid global result computation as well.
One of the problems in semiconductor production is that the production process requires a lot of switching between machines that process lots one after the other and machines that process batches of lots at once, often with long processing times. In this paper, we address this switching problem with the ``boids'' flocking algorithm that was originally used in robotics and movie industry. The flocking behavior is a bio-inspired algorithm that uses only local information and interaction based on simple heuristics. We show that this algorithm addresses these valid considerations in production plant optimization, as it reacts to the switching of machine kinds similar to how a swarm of flocking animals would react to obstacles in its course.
}


\usepackage{cite}
\usepackage{amsmath,amssymb,amsfonts}
\usepackage{algorithmic}
\usepackage{graphicx}
\usepackage{textcomp}
\usepackage{xcolor}

\usepackage{url}
\usepackage{comment}

\begin{document}

\pagestyle{headings}

\maketitle


\section{\uppercase{Introduction}}

Many modern production plants are organized by the job-shop principle, which is a well-known NP-hard problem. This paper is inspired by the use case of the semiconductor manufacturer Infineon Technologies\footnote{Infineon Technologies, \url{https://www.infineon.com/}} which manufactures integrated circuits in the logic and power sector. Their plant produces 1500 products each in around 300 process steps by using up to 1200 stations~\cite{schranz2021bottom_swilt,khatmi2019swarm}. At this problem size, linear optimization methods cannot be applied to compute global plant optimization in a reasonable amount of time~\cite{lawler1993sequencing}. 
Besides linear optimization, swarm intelligence algorithms that are centrally computed have been applied to job-shop scheduling as well. Although they show good performance, they have the same limitations in computation time and complexity as linear optimization methods~\cite{khatmi2019swarm}. For more details the reader is referred to Gao et al.~\cite{gao2019review} for a comprehensive review. To overcome these limiting conditions,\cite{schranz2021bottom_swilt} proposed the novel approach to model the production plant as an agent-based system, where a swarm intelligence algorithm acts on local rules, local knowledge and local interactions. This shifts the problem of computing a global scheduling solution to the design of a distributed algorithm that produces an emergent scheduler from the bottom-up. Such an optimization approach can dynamically react and adapt to changes, like, e.g. tool downs.

In this paper, we use the flocking behavior of agents, so-called ``boids'' as proposed by Reynolds~\cite{reynolds1987flocks}, as an innovative approach to inspire a job-shop scheduling algorithm from the bottom-up. This algorithm---originally used in robotics and the movie industry---inspired a swarm algorithm for production plants in which machines that work on one product (or lot) after the other (single-step machines) alternate with machines that process a whole batch of lots. Typically, only lots of the same type may get processed together in a batch machine. In this case, the problem arises that single-step machines can starve subsequent batch machines from filling their batch efficiently if all the lots of the same type are scheduled to go through the same single-step machine that is upstream in the production process. We hypothesized in previous work \cite{umlauft2023bottom} 
that switching between single-step and batch machines is one of the factors that make optimizing these kinds of production processes especially hard.

In this paper we introduce a model of the production plant and the problem to optimize in Section~\ref{sec:model}. In Section~\ref{sec:algorithm}, we describe the general idea of the flocking model as derived from literature and the adaptation of this idea to our factory optimization problem.
We evaluate the resulting algorithm in the NetLogo-based SwarmFabSim simulation framework, comparing its performance to an engineered algorithm called ``Baseline'' that uses FIFO queues and a fill-the-least-empty-batch-first approach (Section~\ref{sec:eval}). We give a more detailed overview of the related work in Section~\ref{sec:related_work}. The paper is closed out with Section~\ref{sec:conclusion} -- Conclusion and Future Work.

\section{\uppercase {Related Work}}
\label{sec:related_work}

When Reynolds published the boids model~\cite{reynolds1987flocks}, the movie industry had a huge interest in using it for computer simulated flocks. This includes the flocking bats in Tim Burton's Batman Returns in 1992 and the wildebeest stampede in Disney's movie The Lion King in 1994~\cite{terzopoulos1999artificial}. Several works in computer graphics and animation followed, including comparative analysis for large scale flocks in virtual reality~\cite{sung2018comparison}, generation of fish swarms~\cite{pilco2019procedural}, or flocking-based strategy games~\cite{jergeus2020flock}.

Besides the movie industry, the steering behavior of the flocking algorithm has been successfully applied in several swarm robotic applications. Prominent examples cover unmanned aerial vehicles (UAVs) that perform collective following, different formation flights, and flocking in a confined area. The authors in Welsby et al.~\cite{welsby2001} used three flying robots built with helium-balloons to prove the concept and showcase the rules of flocking. De Nardi et al.~\cite{de2006swarmav} presented a flock of helicopters. In the work of Hauert et al.~\cite{hauert2011reynolds} they showcased an autonomous flock of 10 fixed-wing UAVs. To realize this experiment, they flew at relatively large distances and fixed, but different altitudes to prevent crashes. An incredible flock of 20 mini UAVs was presented by Kushleyev et al.~\cite{kushleyev2013towards}. Nevertheless, navigational instructions were calculated on a central workstation. Additionally, Varga et al.~\cite{varga2015distributed} presented a flock of fixed-wing micro-aerial vehicles (MAVs) that is based on flocking-kind of behavior with attraction and repulsion. It was the first flock where the processing relied on local information only. Further experiments were continued~\cite{vasarhelyi2018optimized,balazs2020adaptive}.
Generally, swarm robotics has various applications of the boids model, either by focusing on path planning of a mobile robot that navigates in dynamic 2D-environments~\cite{espelosin2013path}, resilient formation control for mobile robot teams~\cite{saulnier2017}, private flocking controllers to hide the identity of the leading robot~\cite{zheng2020}, or flocking and formation of a group of nonholonomic wheeled mobile robots~\cite{pliego2023flocking}, just to name a few aspects.  
In the application domains of swarm robotics and computer graphics, gaming, and animation, the Reynold's flocking behavior gained huge interest, use, and implementation. 

Algorithms that are inspired by/similar to flocking, such as particle swarm optimization \cite{Kennedy1995PSO}, or migrating birds optimization \cite{Duman2012MBO,Gao2016mbo-for-FJSP,Sioud2018enhancedMBO} and others \cite{Barve2013overview} have been applied to the job-shop scheduling problem before, but only in a centrally calculated way where the algorithm is applied to solutions in the solution space which suffers from the problem of calculation overhead for very large problem sizes.
To the best of our knowledge, no one has so far tried to apply the flocking algorithm in the job-shop scheduling domain as a literal algorithm from the bottom up which makes this research work highly innovative. We model the lots to be produced as boids which ``fly'' through the machine park to achieve an emergent schedule from the bottom-up based on local decision making and interactions.

\section{\uppercase{Production Plant and Simulator Model}}
\label{sec:model}

Our (semiconductor) production plant model consists of several workcenters $W^m$ which contain machines $M^m$ of the same type $m$ each. In our abstraction, a machine of type $m$ can perform exactly one specific process step $P^m$.
The products (in the semiconductor industry, these are so-called lots of 25 wavers each) have a type $t$ that is defined by the recipe $R^t$ of the lot $L^t$. The recipe defines the necessary process steps in the correct order to produce a lot: $R^t = \left[P^m_1, P^m_2,...\right]$.
We use a scheduling model where every machine $M_i$ in the system uses its own, dedicated queue $Q_i$.
Each machine type belongs to one of two kinds: single-step or batch-oriented. Single-step machines process one lot after the other. Batch machines have a batch size \emph{BS} (which depends on their machine type $m$) and can process a whole batch of \emph{BS} lots at once. These lots have to be of the same lot type $t$, though, to be able to be processed together in one batch at the same time (iow. mixing lot types in a batch is prohibited).

Batch machines are typically machines which have a longer processing time than single-step machines. Therefore, in the optimal case, batch machines would only start when their batch is completely filled. This depends on the current availability of enough lots of the same type $t$ to fill the batch. To enable a batch machine to start eventually, even if never enough lots of the same type arrive in its queue, in our model, batch machines employ a timeout-timer \emph{WT}. When \emph{WT} expires, a batch machine can start with a semi-filled batch.

One problem arises from the fact that over the course of the production process lots need to be processed by single-step and batch machines in an alternating fashion. This is because the kind of machine depends on the technical process (e.g., furnaces with long run-times are typically batch machines).
When lots are processed in single-step machines, they leave the machine one by one but a subsequent batch machine expects a whole batch of size \emph{BS} of lots of the same type $t$. Therefore, if all necessary lots of the same type $t$ are scheduled through the same preceding single-step machine, they will be held up in this machine's queue and starve the batch machine from filling its batch in a timely fashion. The batch machine then will typically experience \emph{WT} timeout and start with an only semi-filled batch.
After a batch machine finishes, it releases the whole batch of lots towards the subsequent machines. For single-step machines, theory~\cite{stidham2002analysis} shows that this kind of WIP (work in progress) wave is not optimal as uniform arrival times are better suited for single-step machines than huge waves of product.
We designed our flocking-inspired algorithm described in Sect.~\ref{sec:algorithm} to tackle this exact problem.

From an agent-based swarm intelligence point of view, we model the lots as active agents. These have several decision points during which they can/must act:
\begin{itemize}
    \item Every lot $L^t_i$ that is not currently processing (or finished) must choose a queue $Q^m_j$ for a machine $M^m_j$ that corresponds to the next process step $P^m$ in its recipe $R^t$.
    \item When a machine $M^m_j$ is ready to process the next lot, it triggers the lots to re-shuffle in its queue $Q^m_j$ according to the flocking rules.
\end{itemize}

We aim to optimize the production in this model with the main focus on overall makespan (how long does it take to produce all lots) and on flow factor. Flow factor is defined as $\sum (qt + rpt) / \sum rpt$ with $qt$ being the queueing time and $rpt$ the theoretical raw processing time and averaged over all lots. We do \emph{not} model setup/loading times or transport times between machines.

\subsection{Simulator Model -- Framework}

We implemented our Flocking algorithm in the simulation framework ``SwarmFabSim'' which is realized in the freely available agent-based simulator NetLogo\cite{wilensky1999netlogo}. SwarmFabSim models the machines, the corresponding queues, and the lots to be produced, as swarm agents that can take active decisions. It supports plugging-in and choosing between several swarm algorithms which are de-coupled from the main simulator code via a call-back API.
A detailed description of the simulation framework can be found in~\cite{simultech22}
; the source code 
can be obtained from our GitHub repository~\footnote{\url{https://swarmfabsim.github.io}.}. 

\subsection{Simulator Model -- The Baseline Algorithm}
\label{sec:baseline}

Our comparison algorithm is an engineered algorithm called ``Baseline'' which is a stateless algorithm that uses only local information to take its decisions. This is deliberate as the design goal of this algorithm was to use it to test our swarm algorithms---which also use mainly local information---against it and because we want to employ our swarm algorithms on problem sizes where a global solution is infeasible to calculate in reasonable time. 

For single-step machines, Baseline assigns lots to the machine with the shortest queue in the workcenter and operates the queue as FIFO.

For batch machines, Baseline looks for semi-filled batches in the queues of all the batch machines of a workcenter and fills the batches with the least missing lots first so that batch machines do not encounter timeout of their \emph{WT} timer and do not start with an only semi-filled batch. For lots with a type for which there is no semi-filled batch in any queue in the workcenter, Baseline chooses the batch machine with the shortest overall queue.
The queues of batch machines are separated by lot type as ``mixed type batches'' can not be processed together; iow. they are ``multi-queues''. For example the queue of a currently busy batch machine might contain a half-filled batch of lots of type $t_1$ and a full batch of lots of type $t_2$. When a batch machine needs to choose which lots to take from its multi-queue, 
Baseline will choose a full batch if one is available (or randomly choose one if there are several full batches waiting). If there is no full batch in the multi-queue, the machine waits until the \emph{WT} timer times out. If a batch fills up during \emph{WT}, that batch is processed immediately. If the \emph{WT} timer reaches timeout and no batch has filled up, the fullest batch is chosen. In case of several batches with the same fullness, one is chosen at random.

\section{\uppercase {The Flocking Algorithm}}
\label{sec:algorithm}

The flocking model~\cite{reynolds1987flocks, reynolds2006boids} comprises three local rules that result in a steering behavior of a swarm of agents. An agent is referred to as a ``boid'' and adjusts its course in response to the positions and speeds of its neighboring flockmates:
\begin{enumerate}
\item Separation: Adjust the steering to avoid getting too close to nearby flockmates, preventing overcrowding.
\item Alignment: Modify the steering to align with the collective heading of nearby flockmates, promoting a synchronized direction.
\item Cohesion: Adapt the steering to move closer to the average position of nearby flockmates, encouraging group cohesion.
\end{enumerate}
Each boid only responds to flockmates situated within a specific, limited vicinity around itself. This neighborhood is defined by both a distance, measured from the boid's center, and an angle relative to the boid's flight direction. Any flockmates positioned beyond this localized neighborhood are ignored. An important characteristic of the flocking behavior is its inherent unpredictability: you can never forecast the direction the flock is moving\footnote{If the reader is interested, we recommend the NetLogo library model of flocking to experiment with this characteristic): \url{https://ccl.northwestern.edu/netlogo/models/Flocking}}. 

The original ``boids'' behavior was extended with both predictive obstacle avoidance and goal-seeking capabilities in \cite{reynolds2006boids}. Predictive obstacle avoidance enables the boids to navigate through simulated environments, avoiding stationary objects. In computer animation applications, a secondary, lower-priority goal-seeking behavior guides the flock to adhere to a predefined path.

\subsection{Bottom-up Flocking in Job-Shop Scheduling}
We were inspired to adapt the flocking principle to our production plant optimization problem by observing the problems described previously in Section~\ref{sec:model}---the switch between single-step and batch machines which disrupts the optimal flow and starves the batch machines if products of the same type get stuck in a queue at the same single-step machine or creates WIP waves after batch machines that can overwhelm subsequent single-step machines.

In our adaptation, the assumption is therefore as follows: In an optimal case, lots of the same product type would be processed by parallel single-step machines at the same time so that a batch at a subsequent batch machine could be readily filled. Conversely, the WIP wave that occurs after a batch is released from a batch machine would be spread over as many subsequent single-step machines as possible. In other words, the ``swarm'' of lots would ``spread out'' when going through a workcenter with single-step machines and ``squeeze together'' at a batch machine similar to how a flock of birds flies through, say, a small gap between houses. 

Analogous to the original flocking model described above, we define:
\begin{enumerate}
\item Separation as the spreading out over multiple single-step machines to prevent overwhelming any individual machine by a WIP wave, and
\item Cohesion as a) gathering lots of the same type at a batch machine to efficiently fill a batch and b) manipulating the order of lots of the same type in the queues of single-step machines so they cohere in time.
\item Alignment is strictly given by the recipe, we do not need to derive a synchronized direction from other flockmates. The lots' recipes can be seen as analogous to an enforced version of the goal seeking behavior described in ~\cite{reynolds2006boids}. Therefore, since the direction of flow is not free but enforced, alignment in the true sense is not necessary in our case.
\end{enumerate}

\begin{figure}[htbp]
    \begin{center}
		\includegraphics[width=1\columnwidth]{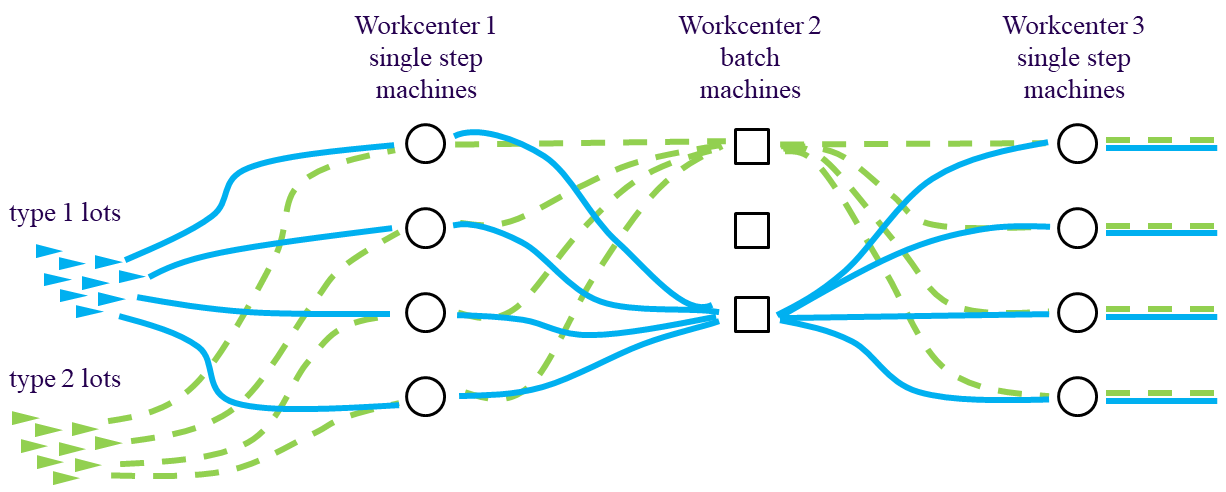}
    \caption{Flocking algorithm concept in a production plant with different kinds of machines; circles representing single-step machines, squares representing batch machines.}
    \label{fig:flocking-concept}    
    \end{center}
\end{figure}

Figure~\ref{fig:flocking-concept} shows the general principle for two flocks of lots with two different types going through three workcenters. In this example, the lot types share these three steps in their recipe; they each need to first be processed by a single-step machine in Workcenter~1, then by a batch machine in Workcenter~2, and finally by a single-step machine in Workcenter~3.

As can be seen, both flocks first spread out over all available single-step machines in parallel. By cohering in the queue, all lots of a type should finish processing at the first workcenter at around the same time and be readily available to fill a batch in a subsequent batch machine at Workcenter~2. The lots of the second type then also arrive closely together in time and fill a batch at a different batch machine at Workcenter~2. When these batch machines eventually finish processing and release their respective lots, these again spread out over the subsequent single-step machines in Workcenter~3.

We engineer the flocking algorithm as a literal swarm intelligence algorithm; instead of applying it in abstracted form to solutions in the solution space, we employ a bottom-up approach where every ``boid'' in the flock corresponds to a literal lot and the global behavior emerges from the local rules and behavior of these lots in the modeled fab.
Our implementation of the Flocking algorithm is realized as follows: 
\begin{itemize}
    \item The lots of every lot type constitute a flock of their own. Iow. only lots of the same type (eg. blue and green color as shown in Fig.~\ref{fig:flocking-concept}) flock together.
    \item The algorithm is implemented at two decision points: When a lot chooses a queue and just before a machine takes from the queue. \item Given the definitions we chose for separation and cohesion, the algorithm must act differently depending on the kind of machine---single-step or batch.
\end{itemize}

At \emph{batch machines}, the flock of lots shall cohere in space and time; iow. lots of the same type shall all go through the same batch machine together. After closer analysis, we find that the Baseline algorithm described above already implements this behavior in an efficient way. We therefore chose to re-use this behavior for the flocking algorithm without changes.

At \emph{single-step machines}, the flock of lots shall spread out over as many machines in the workcenter as possible. Therefore, when a lot needs to decide which machine queue to choose, 
it inspects all the queues in the given workcenter. Lots of the same type in a queue ``discourage'' new lots of this type from choosing this queue, following the rule of separation. As typically queues will not be completely devoid of lots of the same type (especially for large numbers of lots), the rule of separation does not forbid the lot from choosing a queue but rather nudges it to choose from the queues with the least amount of lots of the same type. Out of these, it then chooses the shortest possible queue, following the common ``shortest queue'' practice.

\begin{figure}[htbp]
    \begin{center}
		\includegraphics[width=0.8\columnwidth]{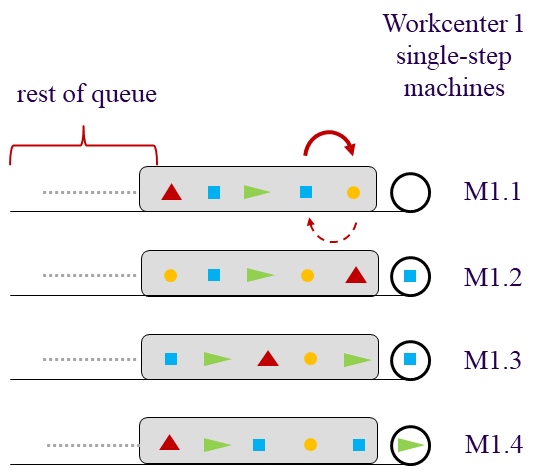}
    \caption{Flock coherence of lots in queues of single-step machines.}
    \label{fig:queue-shuffle}    
    \end{center}
\end{figure}

When a machine is ready to take in a new lot, it triggers the rule of cohesion in time. The lots in the queue then reshuffle as shown in Fig.~\ref{fig:queue-shuffle}:
As queues can grow very long, potentially with several hundred lots per queue, we only consider a subset of the queue that is close to the machine. These first lots then reshuffle to cohere depending on how the lots at all other machines and queues in the same workcenter are distributed. This corresponds to the limited vicinity/visibility principle of the boids algorithm in that lots do not look at the complete queue but only at a short subqueue near the machine.

\subsection{Exemplary Application}
Given the example of Fig.~\ref{fig:queue-shuffle}, the machine about to take a new lot from its queue is M1.1. 
Every lot in the shaded part of the queue (first lots queue, FLSQ) now compares its distance from the machine with the distance of the first lot of same color in all other FLSQs and machines in the workcenter. If another machine is already processing a lot of this type, its distance counts as $0$ (e.g., the square blue lots in M1.2 and M1.3). If it finds a lot of the same type in the first place of the queue (e.g., the square blue lot in the queue of M1.4), this counts as a distance of $1$, the second place in the queue counts as a distance of $2$, etc.
Taking the example of M1.1 and its queue, 
the circular orange lot in our own queue would find that it has a distance of $1$ to the machine, while the average distance of the first circular orange lots in all other FLSQs in this workcenter is equal to $2$. While the square blue lot in the queue of M1.1 finds that the average distance of the other first square blue lots is one third, compared to its own distance of $2$.

Whenever the distance of a lot under consideration is larger than the average distance of the first lots of the same type in the other FLSQs, it attaches a ``pull'' of $-1$ to itself. If the current lot's distance is smaller than that of the others, it attaches a ``pull'' of $+1$. If it dos not find a lot with the same type in any of the other FLSQs, it attaches a neutral ``pull'' of $0$.

After calculating all pull values, the lots re-order themselves in the FLSQ. Lots move $1$ place towards the head of the queue if the pull is $-1$ and $1$ place towards the back of the queue if the pull is $+1$. Lots with a neutral pull do not move. All moves are capped by $[1...length(FLSQ)]$ with $1$ being the head (closest place to the machine) in the FLSQ. Lots perform these movements in random order which results in ``fuzzy'' instead of exact movements because other lot movements that insert a lot in front can affect the position of a lot that is moved later. Like in the boids model, this leads to a bit of unpredictability in the lot movements.

After the lots have re-ordered in the queue, the machine (in our example M1.1) takes the lot that now occupies the head of its queue.

Note that only the front of the queue of the machine that is actively taking a lot is re-ordered. The lots in the other queues re-order themselves when their respective machine wants to take the next lot. 
As the lot order in the queue is only relevant when a machine is about to take a lot and all machines in a workcenter have the same processing time, this does not change the behavior of the algorithm.

\section{\uppercase {Evaluation and Results}}
\label{sec:eval}

\subsection{Simulation Settings}

We evaluated the algorithms in the Small Fab scenario described below. This scenario consists of 5 different workcenters with machine types ($m={0 ... 4}$) to focus on the problem of switching between single-step and batch machines. The parameters for the scenario are shown in Tables~\ref{tab:scenarioparameters} and \ref{tab:machineparameters}.

\begin{table}[!htb]
\caption{Overall parameters of the Small Fab scenario.}
\label{tab:scenarioparameters}
    \begin{center}
    \begin{tabular}{@{}llll@{}}
    \hline
        Parameter &  \\ \hline
        Machine types & 5 \\ 
        Machines / type & $2 ... 6$ \\ 
        Product types & 10 \\
        Recipe length & 16 \\ 
        Lots per type & $6 ... 15$ \\ \hline
    \end{tabular}        
    \end{center}
\end{table}

Out of these 5 machine types, type $m=2$ is a batch type, while the other machines are single-step types. The batch machines have a batch size of 4 lots each and a timeout timer of $\mbox{\emph{WT}}=0.3$~hrs which equals $1/5$ of the raw processing time of 1.5~hrs. 
For simplicity, all single-step machines types have a raw processing time of 0.2~hrs.

\begin{table}[!htb]
\caption{Machine parameters used in the simulation.}
\label{tab:machineparameters}    
    \begin{center}
    \begin{tabular}{@{}llllll@{}}
    \hline
        Machine Parameter  & m=0 & m=1 & m=2 & m=3 & m=4 \\ \hline
        \\[-1em] 
        Number of machines & 5   & 4   & 6   & 2   & 2 \\
        Raw process time [hrs]   & 0.2 & 0.2 & 1.5 & 0.2 & 0.2\\
        Batch size         & 1   & 1   & 4   &   1 &  1\\
        Waiting time [hrs]      & -   & -   & 0.3 &  -  & - \\ \hline
    \end{tabular}
    \end{center}
\end{table}

\begin{figure}[htbp]
    \begin{center}
		\includegraphics[width=0.8\columnwidth]{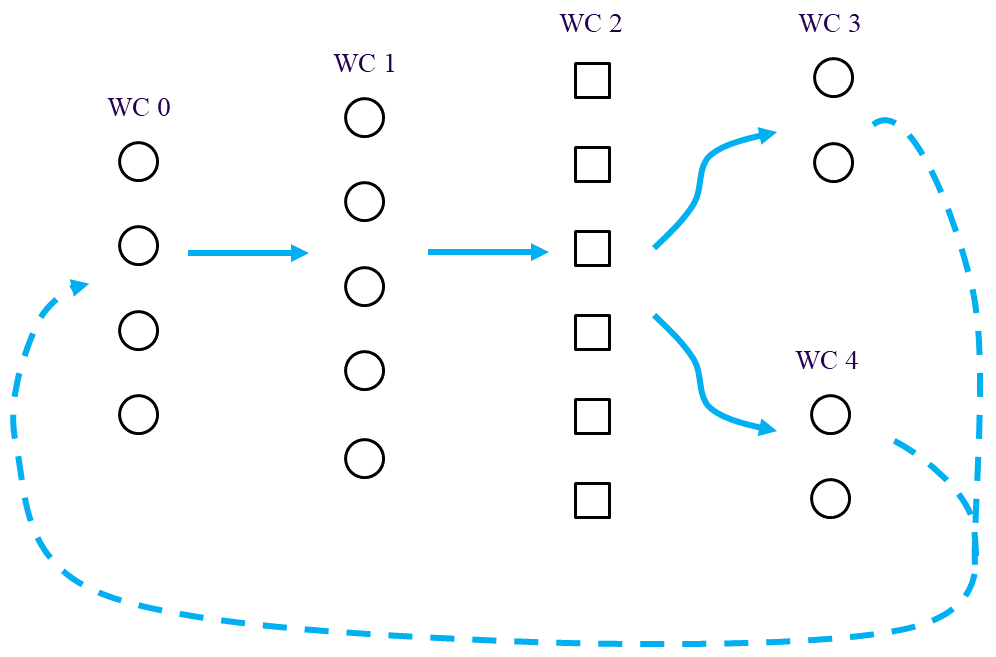}S
    \caption{Small Fab scenario, circles symbolize single-step machines, squares symbolize batch machines.}
    \label{fig:smallfab}    
    \end{center}
\end{figure}

The setup is shown in Fig.~\ref{fig:smallfab}. The workcenter numbers correspond to the respective machine types; iow. WC~0 consists of machines of type $m=0$, WC~1 consists of machines of type $m=1$, etc.

There are lots of 10 different product types in this scenario. This number is inspired by the number of different products in the SMT2020 \cite{SMT2020} testbed for low volume/high mix production (SMT2020, dataset~2), a testbed/dataset widely used in the semiconductor production simulation community.

The number of lots to be used was calibrated to result in a flow factor of around 3 for the Baseline algorithm; the different lot types have between $6...15$ lots as per Table~\ref{tab:scenarioparameters} for a total of 105 lots. Semiconductors are manufactured by depositing several layers onto a silicone wafer; for each layer, the same types of process steps have to be performed with slight variations. In our Small Fab model, all lots share a part of the recipe---they all need to be first processed in WC~0, then in WC~1, then in WC~2, but recipes diverge between WC~3 and WC~4 depending on lot type and layer. After a layer is finished and a lot is released from a machine in either WC~3 or WC~4, it returns to WC~0 for the application of the next layer until the lot is finished. This small scenario produces lots with 4 layers which results in a recipe length of $(4\mbox{ layers}) * (4\mbox{ workcenters in the path)} = 16 \mbox{ steps}$.

The lots are introduced all at once at the entrance to the scenario (the queues of WC~0); we do not model introduction over an extended time-frame (unlike in a real-world fab, where lots will be ordered by customers and introduced into the fab over time, spaced weeks or even months apart depending on load of the fab). Iow., in our model, the fab is initially empty and the 105 total lots are considered as one ``order'' that is introduced at once.

\subsection{Results and Discussion}

We averaged over 50 simulation runs for the Baseline and Flocking algorithm each.
Both algorithms are evaluated according to the following performance metrics: makespan, average flow factor, average tardiness, and average machine utilization. We use the following definitions of these metrics:

\begin{itemize}
    \item \emph{Makespan (MS)} is defined as the time it takes to finish producing all lots.
    \item \emph{Flow factor (FF)} is defined as $\sum (qt + rpt) / \sum rpt$ with $qt$ being the queueing time and $rpt$ the theoretical raw processing time without any queueing delays; averaged over all lots. This indicates by how much of a factor lots are slowed down due to having to wait in queues as opposed to being processed with no waiting times at all.
    \item \emph{Tardiness (TRD)} is usually defined wrt. to an external deadline / due date set by the customer or by factory management. Since our simulator does not model external deadlines, we use a slightly different definition---in our model, tardiness specifies the amount of time the lots spent waiting in queues (averaged over all lots).
    \item \emph{Utilization (UTL)} is calculated as the percentage of time ticks machines are used versus the total number of simulation time ticks (averaged and normalized by the number of machines).
\end{itemize}

The results for these metrics, averaged over 50 simulation runs, are shown in Table~\ref{tab:eval_small}.

\begin{table}[!hbt]
    \caption{Performance comparison for the small Fab Scenario. Changes in \%, positive values denote improvement over Baseline.\\
    MS ... average Makespan, FF ... average Flow Factor,\
    TRD ... average Tardiness, UTL ... average machine utilization
    }
    \label{tab:eval_small}    
    \begin{center}        
    \begin{tabular}{@{}lllr}
    \hline \\[-1em]
         & \small Baseline & \small Flocking & \small Change\\ 
          \\[-1em] \hline
          \\[-1em] \hline
        \small MS [ticks] & \small 324.46 & \small 326.22 & \small -0.54\% \\
        \\[-1em] \hline
        \small FF & \small 3.01 & \small 2.99 & \small 0.55\% \\
        \small TRD [ticks] & \small 168.97 & \small 167.57 & \small 0.83\%  \\ 
        \small UTL [\%] & \small 68.83 & \small 68.45 & \small 0.54\%  \\ \hline
    \end{tabular}
    \end{center}
\end{table}

The performance evaluation shows that the Flocking algorithm performs only marginally better than the Baseline algorithm for average flow factor, tardiness, and utilization, while the average makespan is slightly worse. 

\begin{figure}[htbp]
    \begin{center}
		\includegraphics[width=0.8\columnwidth]{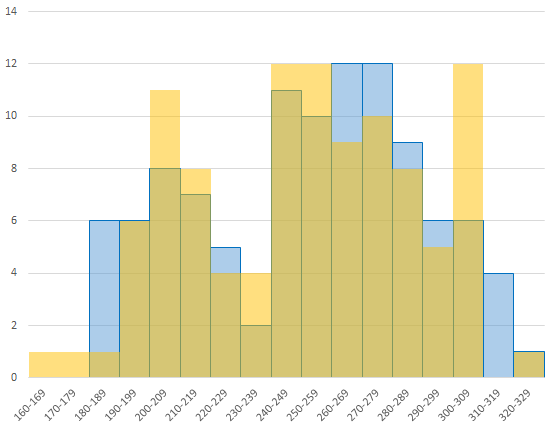}
    \caption{Histogram of lot production finish times, binned every 10 ticks. Baseline algorithm denoted in blue, Flocking algorithm denoted in yellow.}
    \label{fig:histo}  
    \end{center}
\end{figure}

In the following we analyze the finishing times of the produced lots to gain insight how a worse makespan can occur despite a slightly better flow factor. A histogram comparing the lot finishing times for both, the Baseline algorithm and the flocking algorithm is given in Fig.~\ref{fig:histo}. Despite an average lower flow factor, the makespan can increase when a few (or even a single) lot(s) finish later than in the setup with the algorithm with the worse flow factor.
As we can see in Fig.~\ref{fig:histo}, the Flocking algorithm (denoted in yellow) finishes more lots at earlier times than the Baseline algorithm (denoted in blue) but has one extra lot that finishes later, increasing the overall makespan.

\begin{figure}[htbp]
    \begin{center}
		\includegraphics[width=0.6\columnwidth]{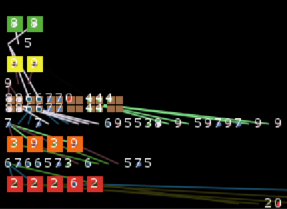}
    \caption{Screenshot of Small Fab as visualized in simulator (detail).}
    \label{fig:screenshot}    
    \end{center}
\end{figure}

When single stepping\footnote{running the simulation manually one tick at a time for demonstration and debugging purposes, not related to single-step machines.} the algorithm in the simulator and observing the behavior of the lots, the desired behavior can clearly be observed; after every call the machines take from queue, the lots visibly cohere more and more. This can also be seen in the visualization of the fab as the parallel machines in a workcenter fill with more and more lots of the same type at the same time. For example, as shown in Fig.~\ref{fig:screenshot}, the 5 red single-step machines of type $m=0$ at the bottom are filled with almost exclusively lots of type $2$. We also see the separation effect; in this example at the 2 topmost green machines, each got loaded with a lot of type $m=8$ after these lots were previously released from one of the 6 brown batch machines (middle).

In comparison, during runs of the Baseline algorithm the lots show a much more random distribution of lot types across queues and in parallel single-step machines in the same workcenter. Despite the correct behavior, the performance of the Flocking algorithm can barely beat the Baseline algorithm, though. In the following, we discuss possible reasons for this:

First, remember that the Baseline algorithm already does the desired lot coherence for batch machines and this function is re-used 1:1 in the Flocking algorithm. Therefore, any difference in performance can only come from the behavior at the single-step machines.

The purpose of spreading the flock over all available single-step machines and cohering them in time is so that subsequent batch machines will not get starved by lots of the same type getting ``stuck'' in one long queue waiting to be processed by a single-step machine upstream in the recipe. Single step machines are typically significantly faster than batch machines, though, and sometimes even faster than a batch machine's \emph{WT} timeout timer. Depending on the number of available machines and processing times, the upstream single-step machines can be able to outperform the subsequent batch machines. Especially if the workcenter with the batch machines is a bottleneck in the fab, there will be enough lots queueing up to fill the batches even when the lots do not explicitly cohere at the upstream machines. Therefore, the performance improvement to be gained by lot coherence in the queues of the single-step machines ist limited.

The final potential source of performance improvement is the separation rule when lots exit a batch machine and get distributed over as many parallel single-step machines in the next workcenter as possible. The effect that the behavior of separating the lots by their type---nudging them to consider queues with the least number of their same type---and subsequently choosing the shortest queue out of these can have is again limited, though, as this is very similar to simply choosing the shortest queue without regard for lot type, especially when large numbers of lots are present.

\section{\uppercase {Conclusion and Future Work}}
\label{sec:conclusion}

This paper has proposed the use of a bottom-up flocking algorithm to address the problems induced by different kinds of machines in semiconductor production plants. Namely, single-step machines that process one lot after the other and batch machines that process a whole batch of lots of the same type at the same time. In front end of line semiconductor production, these (and several other) kinds of machines alternate during the production process, which makes it especially hard to optimize this special case of job-shop problem. In the optimal case, enough lots of the same type arrive at a batch machine to fill up the whole batch so the machine can start quickly and fully filled for optimal utilization.
If, instead, due to unfortunate scheduling, lots of the same type are held up in the queue of a preceding single-step machine, the batch machine will have to either wait until the batch fills up or start with an only semi-filled batch.
In the opposite case, a batch machine eventually releases all its lots at the same time, creating a wave of lots for the subsequent single-step machines. Ideally, these lots would again spread out to use as many machines as possible in parallel to a) neither overwhelm any individual machine and b) create time-synchronized batches ``flocking'' through the production plant for further, subsequent batch machines.

Our adaptation of the flocking principle endeavors to achieve exactly this: Lots of the same type are considered a flock that tries to ``fly'' through the production plant by spreading out across as many machines as possible in workcenters with single-step machines while cohering in time and space at batch machines.

We apply the flocking principle as a literal swarm intelligence algorithm in a bottom-up manner, where the lots are considered active agents. Agents only use information from their current local neighborhood (workcenter) to take decisions from which the global solution then emerges automatically.
We have used NetLogo to simulate the production plant and demonstrate the feasibility of the principle. The performance was measured compared to an engineered baseline algorithm, considering makespan, flow factor, tardiness, and machine utilization as performance indicators.

Our results show that while this type of algorithm shows only a small performance improvement in the tested setting, this is achieved using only low-effort local calculations. A flocking approach may still be beneficial when used in combination with other algorithmic behavior in front of a large workcenter of batch machines which might run the risk of starvation by a bottleneck workcenter of single-step machines that is located ``upstream'' for most of the recipes of the lots being processed.

Future work will investigate the proposed algorithm using larger datasets with higher levels of realism, such as the SMT2020 dataset commonly used in detailed semiconductor production simulation.

\section*{\uppercase {Acknowledgement}}
This work was performed in the course of project SwarmIn supported by FFG under contract number 894072.

\end{document}